\documentclass{fairmeta}

\usepackage{amsmath}
\usepackage{amssymb}
\usepackage{array}
\usepackage{tabularx}

\newcommand{\pio}{$\pi_{0.5}$}
\newcommand{\NGM}{NGM}
\newcommand{\tabnote}[2]{\par\vspace{3pt}\begin{minipage}{#1}\footnotesize #2\end{minipage}}

\title{Fast Plans, Faithful Actions:\\ Closing the Planning--Execution Gap in Hierarchical Vision-Language-Action Models}

\author[1,*,\dagger]{Chuanliang Xie}
\author[1,*]{Boyu Ma}
\author[1]{Gen Li}
\author[1]{Yizhou Liu}
\author[1]{Houwang Chen}
\author[1]{Xinyu Zhou}
\author[1,\dagger]{Jianfei Yang}

\affiliation[1]{MARS Lab, Nanyang Technological University}

\contribution[*]{Equal contribution}
\contribution[\dagger]{Corresponding authors}

\abstract{
Hierarchical vision-language-action (VLA) systems consist of a high-level vision-language planner and a low-level action expert that generates continuous actions.
This hierarchical design has practical value only if the planner can generate plans fast enough to meet real-time control requirements, and the resulting plans actually contribute to the generation of action.
We study one such system, a waypoint hierarchy pipeline adapted from $\pi_{0.5}$, and find that neither requirement is satisfied.
This baseline relies on token-level autoregressive decoding (Token-AR) to generate a waypoint plan, requiring 57 very expensive vision-language model (VLM) forward passes.
However, we find that erasing the waypoint endpoints has little effect on task success.
Two findings reveal the misalignment of planner--executor: the planner generates outputs at an excessively fine granularity, and the executor underuses plans as a control condition.
We address the latency issue with waypoint-aligned block-autoregressive decoding (Block-AR), and plan underuse issue with normalized goal modulation (NGM), a layer-wise goal path constrained by phase gating and anti-shortcut training so that the waypoint influences action generation maintaining other signals.
Our method reduces the maximum number of VLM forward passes from 57 to 8 on LIBERO, including one prefix prefill, and achieves an $8.7\times$ reduction in planning latency on a Rokae dual-arm robot.
With normalized goal modulation and anti-shortcut training, Block-AR's success rate on LIBERO-Long increases from 91.0\% to 96.2\%, while its average success rate across the four suites increases from 95.85\% to 98.45\%.
On three bimanual tasks with this robot, success rates remain comparable across methods.
}

\correspondence{\email{jianfei.yang@ntu.edu.sg}, \email{chuanlia001@e.ntu.edu.sg}}

\begin{document}

\maketitle

\section{Introduction}

Vision-language-action (VLA) models generate robot actions directly from image and language inputs~\citep{openvla,pi0,pi05}. In recent years, more and more works have adopted a hierarchical structure to realize this mapping: a high-level vision-language planner first generates a plan in an intermediate form, such as a trajectory sketch, a set of keyposes, or a waypoint sequence, and then an execution stage converts this plan into continuous actions~\citep{hamster,hamster3d,gae,notvla,coarse2control}. The benefit of this division of labor is that a large vision-language model (VLM) can fully exploit its advantages in semantic and spatial reasoning, and can incorporate action-space chain-of-thought~\citep{coarse2control}, while the execution stage handles continuous control.

\begin{figure}[t]
\centering
\includegraphics[width=\textwidth]{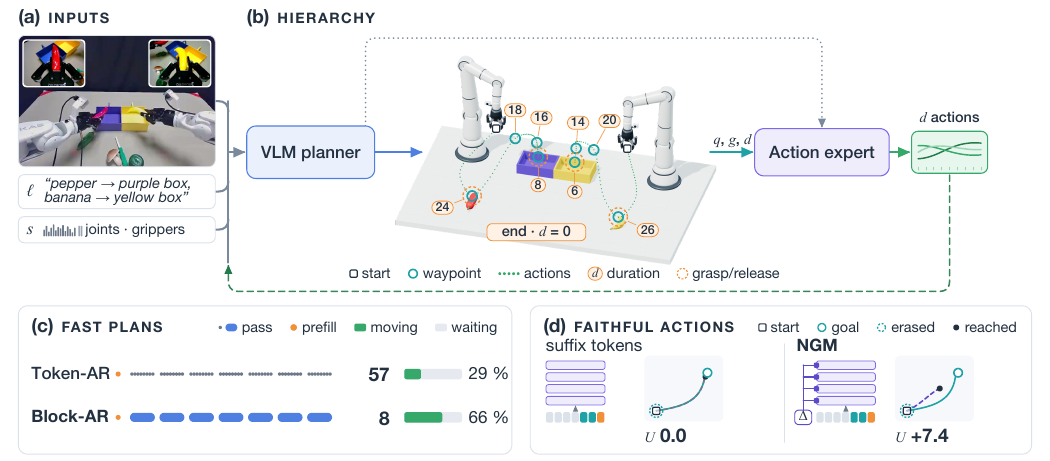}
\caption{Fast plans, faithful actions. (a)~Inputs (Rokae AR5-5 dual arm, pepper--banana placement): external and wrist images, instruction $\ell$, state $s$. (b)~A PaliGemma planner (Gemma-2B language model) predicts $(q,g,d)$ waypoints (target configuration, gripper, duration). The schematic shows four waypoints per arm (hollow rings; orange dashed rings: grasp/release), illustrative per-arm step counts (linked pills), and corresponding action beads. Deployed dual-arm blocks share one $d$; $d=0$ ends the plan. A 300M flow-matching expert, which also attends to the planner's image and language prefix (dotted), maps each waypoint to $d$ actions; the robot executes, observes, and replans (dashed loop). (c)~For one seven-waypoint LIBERO plan, Token-AR needs one prefix prefill plus one backbone pass per value token (dots), 57 passes in total; Block-AR needs the same prefill plus one pass per waypoint (pills), 8 in total. On the robot, plan latency falls from 1094 to 125\,ms and the arms move 66\% instead of 29\% of the time (bars). (d)~Endpoint-erasure test on LIBERO-Long: with images, instruction, state, and flow noise fixed, the commanded endpoint is set to the segment start. Erasure does not reduce success with suffix-token conditioning ($U=0$), but reduces success by 7.4 points with normalized goal modulation (NGM), which injects the goal into every expert layer.}
\label{fig:teaser}
\end{figure}

However, this division of labor places two requirements on the interface between the planner and the executor. Requirement 1 (R1): the plan must be generated in time. When the VLM generates the plan token by token in an autoregressive manner, the cost of decode computation grows sharply with the number of output tokens, which poses a great challenge to the real-time performance of the whole system. Requirement 2 (R2): the plan must genuinely contribute to behavior. Besides the plan, the low-level policy can usually also see the key/value (K/V) representations of the images and language; in this case, empirical risk minimization does not force it to determine its control behavior through this intermediate representation. Prior work has analyzed failures in which a learned executor deviates from the path predicted by the planner~\citep{hamster}; this paper goes one step further and directly measures this interface through controlled interventions. Fig.~\ref{fig:teaser} outlines the hierarchical architecture we study and the two requirements above.

To study these two requirements, we adapt $\pi_{0.5}$ into a hierarchical architecture, which we call Waypoint Token-AR (Fig.~\ref{fig:teaser}(a), (b)). In this architecture, the PaliGemma backbone~\citep{paligemma} serves as the planner and is responsible for predicting waypoints in the robot's configuration space; each waypoint consists of three parts: a target configuration, a gripper state, and an execution duration. The action expert trained with flow matching~\citep{flowmatching} receives the current state and the waypoint through suffix tokens; we call this baseline route suffix-token conditioning. It should be noted that this baseline is built by us specifically for controlled experiments and is not the original architecture of $\pi_{0.5}$.

We find that neither requirement is satisfied on this baseline. Consider R1 first: generating a LIBERO~\citep{libero} plan containing seven waypoints requires the 2B-parameter language backbone to perform 57 serial backbone passes, including one prefix prefill. On a real dual-arm robot, the time spent generating one plan even exceeds the execution time of the action segment that the plan specifies; as a result, the arms are in a waiting state for 71\% of the actual running time. Now consider R2: with the images, language, and flow-matching noise all held fixed, we erase the target endpoint in the plan, and the resulting change in actions is only 0.6\% of the change caused by re-sampling the noise once. Erasing the target endpoint during online execution has little effect on task success. By contrast, the action expert does show a measurable response to the execution duration.

These two findings demonstrate a misalignment between the planner and the executor.
The executor consumes a plan one waypoint at a time, but the planner generates it one token at a time.
The waypoint-aligned block-autoregressive decoding (Block-AR) removes this granularity mismatch: the planner generates a complete waypoint in each VLM forward pass while the causal dependency between waypoints can be preserved (Fig.~\ref{fig:teaser}(c)).
The executor takes in the waypoint as a condition, but the waypoint contributes little to action generation.
To solve this issue, we design normalized goal modulation (NGM), which gives the goal a layer-wise path into the action expert (Fig.~\ref{fig:teaser}(d)).
We explain why an unconstrained path produces a ``displacement-integrator shortcut'' and then constrain it with phase gating and anti-shortcut training, thereby ensuring that waypoint information contributes to action generation while preserving the model's responses to other input signals.

The main contributions of this paper are as follows:

\begin{itemize}
\item We conduct a controlled study on a waypoint hierarchy built on $\pi_{0.5}$. By measuring action sensitivity and task utility under waypoint interventions, we find that standard suffix-token conditioning leaves the spatial target endpoint almost ineffective.
\item We propose waypoint-aligned block-autoregressive planning. It reduces the maximum serial planning depth from 57 forward passes to 8, and in experiments reduces the generation latency per plan from 1094\,ms to 125\,ms, raises the proportion of motion time from 29\% to 66\%, and supports a learnable end-of-plan marker.
\item We identify the displacement-integrator shortcut that makes direct goal injection fail, and then design normalized goal modulation (NGM) as a layer-wise goal path whose phase gate, condition noise, and dropout mitigate the training shortcut. NGM improves Block-AR performance on all four LIBERO suites while giving the waypoints a real effect on actions.
\end{itemize}

\section{Related Work}

\subsection{Waypoint and keypose interfaces}

Sparse intermediate representations have a long research history in imitation learning, including generating motion between keyposes with diffusion or consistency models~\citep{chaineddiffuser,hdp,bikc}, salient-point hybrid methods~\citep{sphinx}, and Automatic Waypoint Extraction (AWE). AWE selects a minimal waypoint set under the constraint of a piecewise-linear reconstruction-error threshold~\citep{awe}. We extract waypoints offline with AWE under gripper-transition and fixed-horizon constraints (Section~\ref{sec:waypoints}). The usual design of hierarchical VLAs is to let the VLM generate the sparse plan itself, for example: 2-D trajectory sketches~\citep{hamster}, pixel-depth waypoint sequences~\citep{hamster3d}, camera-frame 3-D waypoints executed by a frozen action expert~\citep{gae}, image-space trajectory tokens executed via splines~\citep{notvla}, and coarse-to-fine control~\citep{coarse2control}. Our VLM also writes waypoints, each with a duration and an end flag that give the low-level policy timing and termination information. We study the serial cost of the above hierarchy and the influence of waypoints (Table~\ref{tab:interfaces}).

\begin{table}[t]
\caption{Planner-to-executor designs in hierarchical VLAs.}
\label{tab:interfaces}
\centering\footnotesize
\setlength{\tabcolsep}{4pt}
\begin{tabularx}{\textwidth}{@{}>{\raggedright\arraybackslash}p{3.2cm}*{4}{>{\raggedright\arraybackslash}X}@{}}
\toprule
Method & Plan space & Executor & Serial cost grows with & Interface assessment \\
\midrule
HAMSTER\newline\citep{hamster} & 2-D image path & separate policy & path points & failure attribution \\
\addlinespace[3pt]
3D HAMSTER\newline\citep{hamster3d} & pixel + depth path & separate flow policy & path points & guidance ablation \\
\addlinespace[3pt]
GAE\newline\citep{gae} & camera-frame 3-D points & frozen point-cloud expert & waypoints & goal-noise ablation \\
\addlinespace[3pt]
NoTVLA\newline\citep{notvla} & pixel + depth keyposes & spline + replanning & keyposes & deterministic execution \\
\addlinespace[3pt]
Coarse-to-Control\newline\citep{coarse2control} & coarse action blocks & same model, AR tokens & blocks + actions & qualitative diagnostics \\
\addlinespace[3pt]
Ours & robot configuration + duration & flow expert, shared prefix & value slots (Token-AR) / waypoints (Block-AR) & $S$, $U$ \\
\bottomrule
\end{tabularx}
\tabnote{\textwidth}{Serial cost summarizes the decoding structure, not measured latency across methods. Interface assessment describes the reported evaluation or execution mechanism. 3D HAMSTER compares no, 2-D, and 3-D guidance; GAE varies goal-pose noise during training; HAMSTER attributes failures to trajectory deviation; Coarse-to-Control visualizes attention and decoded plans. Our interventions change the waypoint while holding other inputs fixed.}
\end{table}

\subsection{Decoding efficiency in VLAs}

OpenVLA generates discretized robot actions autoregressively~\citep{openvla}, whereas $\pi_0$ uses a flow-matching action expert to generate continuous action chunks~\citep{pi0}. Serial VLM decoding can limit the control frequency of autoregressive VLAs. Approaches to improving efficiency include frequency-domain action compression in FAST~\citep{fast}, parallel decoding and consistency decoding~\citep{pdvla,ceedvla}, asynchronous reasoning and execution in FiS-VLA~\citep{fisvla}, and joint discrete/continuous training with knowledge insulation for fast action-expert inference~\citep{ki}. Block decoding itself is not a new concept. The difference is that we align each block with one semantically meaningful waypoint, and jointly decode its target configuration, gripper state, duration, and $d=0$ termination flag in a single backbone forward pass; this reduces serial depth and improves termination decoding (Section~\ref{sec:fast}).

\subsection{Conditions ignored by the policy}

Learned policies can rely on shortcuts or underuse relevant conditions. Related failures include causal confusion and copycat behavior in behavior cloning~\citep{causalconfusion,copycat}, and weak language conditioning in VLAs~\citep{langforce,cast}. Guidance can strengthen condition adherence in generative models~\citep{cfg,guidedflows}. Hindsight experience replay, in turn, relabels past experience with goals that were actually achieved~\citep{her}. We encountered a similar problem in our hierarchical architecture: the waypoints generated by the VLM are drowned out by the image and language signals, so that they have almost no influence on the final action generation. To address this problem, we propose closed-loop measurement methods and a corresponding training fix.

\section{Problem Formulation}
\label{sec:formulation}

\textbf{Plan representation.} Let $o$ be the current observation (a first-person image and a wrist-camera image), $\ell$ the language instruction, and $s\in\mathbb{R}^{n}$ the robot proprioceptive state. The planner $\pi_H$ predicts at most $M$ waypoint blocks at a time. The executable plan consists of the $K$ waypoints before an end code appears or the decoding limit is reached:
\begin{equation}
P=\pi_H(o,\ell,s)=\big(w_1,\dots,w_K\big),\qquad w_i=(q_i,g_i,d_i),
\label{eq:plan}
\end{equation}
where $K\le M$; $q_i$ is the discretized target robot configuration; $g_i\in\{\text{open},\text{closed}\}^{n_g}$ contains one gripper command for each arm; and $d_i\in\{1,\dots,D\}$ is the number of continuous actions required to reach $q_i$. A block with $d=0$ ends the decoded plan; task completion is assessed separately.

\textbf{Execution.} The executor $\pi_L$ takes each waypoint in turn as the current target state and generates continuous actions accordingly. Given the current state $s_i$ and the waypoint $w_i$, the executor outputs $d_i$ actions:
\begin{equation}
a_{1:d_i}=\pi_L\big(o,\ell,s_i,\,q_i,g_i,d_i;\,\epsilon\big),
\label{eq:exec}
\end{equation}
where $\epsilon$ is the noise of the generative action expert. When every decoded duration does not exceed the prediction horizon $D$ of the action expert, we call the representation horizon-compatible. This property only guarantees that the requested action segment fits within the output horizon of the action expert; it does not imply that the motion is collision-free or dynamically feasible. Section~\ref{sec:waypoints} guarantees horizon compatibility by construction.

\textbf{Faithfulness.} Even if a waypoint is horizon-compatible, it may still have almost no effect on actual execution. Because the executor also receives image and language signals through K/V representations, the task information carried by $q_i$ may be ignored or drowned out, especially when each training context corresponds to only one endpoint. We assess the model's dependence on waypoints with two controlled intervention experiments. When evaluating sensitivity, we fix $o$, $\ell$, $s_i$, and $\epsilon$ and change only the waypoint. Sensitivity compares the action change caused by a waypoint intervention $\delta$ with the change caused by re-sampling the action expert's own noise:
\begin{equation}
S(\delta)=\frac{\operatorname{med}\|\pi_L(\cdot,q+\delta;\epsilon)-\pi_L(\cdot,q;\epsilon)\|}{\operatorname{med}\|\pi_L(\cdot,q;\epsilon')-\pi_L(\cdot,q;\epsilon)\|},
\label{eq:S}
\end{equation}
where the norm is first computed as the root mean square (RMS) over the executed time steps and valid action dimensions, and then the median is taken over all segments. Utility is the paired change in task success rate after the spatial plan is erased at test time, i.e., $U=\mathrm{SR}(P)-\mathrm{SR}(\tilde P)$, where $\tilde q_i:=s_i$. Thus, $U>0$ indicates that erasing the endpoint hurts task success. Neither of the above measures alone can prove that the conditioning input is effective; we interpret them together with the success rate under the intact plan and the model's sensitivity to visual input. Section~\ref{sec:faith} explains why this joint interpretation is necessary: a model can have a very large $S$ and still fail to complete the task.

\section{Method}

\begin{figure}[t]
\centering
\includegraphics[width=\textwidth]{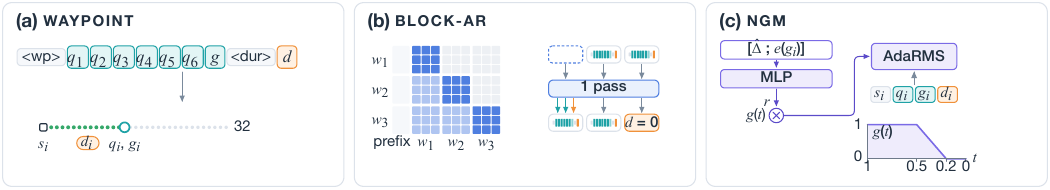}
\caption{Method. (a)~Waypoint representation: a LIBERO waypoint is the token group \texttt{<wp>}$\,q_1\cdots q_6\,g\,$\texttt{<dur>}$\,d$ with $d\in\{1,\dots,32\}$ (338 reserved token IDs; a robot waypoint has 19 tokens). For block $i$ the expert denoises $D=32$ actions from the measured state $s_i$ and executes the first $d_i$ (green); the next block starts from the measured state. (b)~Block-AR: attention is bidirectional within a block (dark) and causal across blocks (light); inputs are shifted by one block (the first is a learned query, dashed) and targets are not, so one backbone pass yields a whole waypoint through three LM-head calls (configuration, gripper, duration). Decoding produces at most $M=7$ blocks and stops early when a block predicts $d=0$. (c)~NGM: besides the suffix tokens $s_i,q_i,g_i,d_i$, the normalized goal $[\widehat\Delta;e(g_i)]$ with $\widehat\Delta=(q_i-s_i)/\sigma_\Delta$ enters an MLP with a zero-initialized output projection. Its residual $r$, gated by $g(t)$, is added to the AdaRMS condition $\phi(t)+g(t)r$ of every layer; the gate is open for $t\ge0.5$ and closed for $t\le0.2$. Training noises the endpoint seen by both routes ($\sigma_c=0.7$) and replaces both by null embeddings with probability 0.15; inference uses the clean endpoint and the gate only.}
\label{fig:method}
\end{figure}

Our system (Fig.~\ref{fig:method}) is derived from $\pi_{0.5}$~\citep{pi05}: a PaliGemma backbone~\citep{paligemma} (consisting of a SigLIP encoder and a Gemma-2B language model) serves as the planner, and a separate 300M-parameter Gemma action expert inherited from $\pi_{0.5}$ serves as the executor and is trained with flow matching~\citep{flowmatching}. The action expert accesses the cached vision and language-instruction prefix through cross-attention, and the two modules are adapted with low-rank adaptation (LoRA)~\citep{lora} within the same coupled model. We add an interface for outputting waypoints to the planner and compare two different ways in which the action expert generates continuous actions conditioned on the waypoints. Success rates and planning costs are reported in Tables~\ref{tab:main} and~\ref{tab:cost}, respectively.

\subsection{Waypoint representation and execution}
\label{sec:waypoints}

\begin{figure}[t]
\centering
\includegraphics[width=0.6\textwidth]{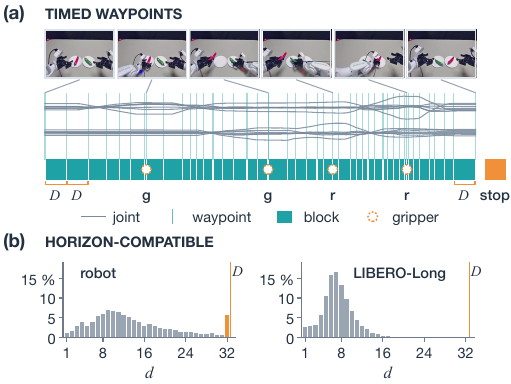}
\caption{Waypoints extracted from demonstrations. (a)~Pepper-swap episode~2: 634 steps (21\,s) become 45 waypoints. The top row shows the start, four gripper transitions, and end. The curves show the 14 joint angles (left arm above, right arm below; centered with a shared scale); vertical cuts mark waypoints, and the blocks below encode segments $(q,g,d)$ with width proportional to duration $d$. Brackets identify the three segments capped at $D=32$ steps; rings mark gripper transitions (g = grasp, r = release; left, right, right, left arm). The orange $d=0$ block marks the end of the plan. (b)~Segment-duration distributions on a shared scale: robot (6{,}158 training segments; median 12, 5.8\% at the cap) and LIBERO-Long (13{,}815 segments; median 7, longest 29).}
\label{fig:waypoints}
\end{figure}

\textbf{Extraction.} We extract waypoints from each demonstration using the dynamic programming algorithm of AWE~\citep{awe}. The algorithm uses a segment-error threshold $\eta$ to select a sparse set of frames. We force the frame at every gripper state transition to be a keyframe and impose a maximum segment-length constraint inside the dynamic program, so that every executable gap falls within $[1,D]$, where $D=32$ is the number of control steps in the action expert's prediction horizon. For the real robot running at 30\,Hz, this corresponds to at most 1.07\,s; for the LIBERO simulator running at 20\,Hz, it corresponds to at most 1.6\,s. The number of steps from the segment start to the target waypoint is set as the positive duration of that waypoint, and a separate terminal block with $d=0$ is added. Fig.~\ref{fig:waypoints} shows an example; segment statistics are in Table~\ref{tab:cost}.

Direct interpolation and inverse kinematics alone do not provide scene-conditioned corrections between waypoints. Following the motivation for a perception-conditioned executor in GAE~\citep{gae}, we use the flow-matching action expert to generate continuous actions between waypoints.

\begin{table}[t]
\caption{Success rate (\%) on the four LIBERO suites.}
\label{tab:main}
\centering\footnotesize
\setlength{\tabcolsep}{5pt}
\begin{tabular*}{\textwidth}{@{\extracolsep{\fill}}l c c c c c c c@{}}
\toprule
Method & Spatial & Object & Goal & Long & Avg. & $\Delta$ vs \pio & $\Delta$ vs Token-AR \\
\midrule
\pio\ (released, as reported) & 98.8 & 98.2 & 98.0 & 92.4 & 96.85 & -- & -- \\
Ours, Token-AR + suffix cond. & 98.0 & 99.0 & 97.6 & 92.2 & 96.70 & $-0.15$ & -- \\
Ours, Block-AR + suffix cond. & 98.6 & 97.6 & 96.2 & 91.0 & 95.85 & $-1.00$ & $-0.85$ \\
Ours, Block-AR + \NGM & 99.2 & 99.4 & 99.0 & 96.2 & 98.45 & $+1.60$ & $+1.75$ \\
\bottomrule
\end{tabular*}
\tabnote{\textwidth}{$N=500$ per suite (10 tasks $\times$ 50 initial states). The \pio\ row lists the values reported in the official OpenPI LIBERO evaluation README~\citep{openpi_libero}; the three waypoint variants use the common protocol of Section~\ref{sec:setup} and share data, initialization, and training steps. $\Delta$: difference of the four-suite averages in points.}
\end{table}

\textbf{Tokenization.} As shown in Fig.~\ref{fig:method}(a), we reuse 338 token IDs at the tail of the existing PaliGemma vocabulary to represent waypoints: 300 token IDs shared by the six continuous configuration dimensions (each dimension is normalized separately before use), two gripper tokens, 34 duration codes, and two structural tokens (\texttt{<wp>}, \texttt{<dur>}). The duration tokens use $d=0$ to indicate end-of-plan and $d\in\{1,\dots,32\}$ to indicate actual durations. Code 33 marks the current-state input and may also arise when training shifts the input observation by $\pm1$ step, adjusting the first duration while keeping waypoint targets fixed. Any decoded 33 is clamped to 32. A LIBERO waypoint is represented by the token group \texttt{<wp>}$\,q_1\cdots q_6\,g\,$\texttt{<dur>}$\,d$; on the dual-arm robot, the token group contains 14 joint slots and two gripper slots. Decoding is slot-constrained: the logits of each slot are masked to its corresponding token family, so every decoded plan is syntactically valid. The planner prefix contains the instruction, the discretized current state, and the camera images (two in LIBERO; one external-view image and one wrist-view image on the real robot). Decoding produces at most $M=7$ blocks; if a terminal block is produced, it is also counted.

\textbf{Execution.} The suffix-conditioned action expert takes the segment start state, the target $(q_i,g_i)$, and the duration $d_i$ as suffix tokens, and denoises an action chunk of length $D$, of which the first $d_i$ actions are executed. The next segment starts from the measured state. In simulation, a new plan is requested only after the current plan has been fully executed; on the real robot, a new plan is requested as soon as the first segment has been executed. The real robot uses receding-horizon replanning because its observation changes faster than a full plan can be executed. The code $d=0$ indicates the end of the current decoded plan; whether the task terminates is determined by the simulator or by a human operator.

\subsection{Waypoint-aligned block-autoregressive planning}
\label{sec:blockar}

\begin{figure}[t]
\centering
\includegraphics[width=\textwidth]{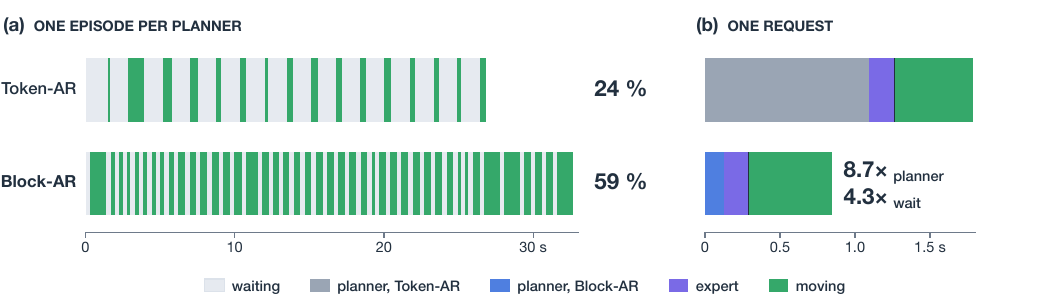}
\caption{Robot planning latency and motion duty cycle. (a)~Recorded shelf-placement timelines for Token-AR and Block-AR on a common time axis. Waiting comprises planner, expert, and transport; motion denotes commanded execution at 30\,Hz. Motion occupies 24\% and 59\% here, versus 29\% and 66\% over 35 telemetry episodes (Table~\ref{tab:cost}). Both episodes were stopped before placement. (b)~Mean request and commanded-motion durations (497 Token-AR / 936 Block-AR requests). Planner latency falls from 1094 to 125\,ms ($8.7\times$); expert latency remains near 160\,ms. Transport is the hairline segment. The request round trip drops from 1266 to 293\,ms ($4.3\times$), below the roughly 0.5\,s of commanded motion.}
\label{fig:fast}
\end{figure}

Compact Token-AR must resolve the value slots sequentially: at the maximum $M=7$, one prefix prefill plus 56 value-token forward passes requires a total of 57 expensive serial backbone forward passes. This becomes the most time-consuming part of the system. To speed up planner inference, we treat each waypoint as one block. Attention remains causal across blocks and is bidirectional within a block; the inputs are shifted by one block rather than one token, so block $k$ is predicted from the preceding blocks $<k$. The first block is predicted from a learnable query with a zero-initialized residual. The labels are not shifted, and both planners use the same training samples and target values. At inference, each decoding forward pass generates all semantic value slots of one waypoint; decoding stops when a block predicts $d=0$. The three token families in a block are read out through three language-model-head calls. Therefore, a maximum-length LIBERO plan needs only 8 forward passes (one prefill plus seven waypoint forward passes) instead of 57. Because the number of Block-AR forward passes is independent of the number of slots per waypoint, the more degrees of freedom the robot has, the more pronounced the speed-up: $57\rightarrow8$ in LIBERO and $120\rightarrow8$ on the robot with 16 degrees of freedom. Fig.~\ref{fig:fast} illustrates the resulting latency savings.

\subsection{Normalized goal modulation}
\label{sec:ngm}

The suffix-conditioned expert receives the goal only through the suffix tokens in \eqref{eq:exec}. Because it can also see the images and language, nothing in the training loss forces it to route task information through the waypoint. We add an additional, more direct route for the goal. We write the goal as a displacement relative to the current state and scale it per dimension: $\widehat\Delta=(q_i-s_i)/\sigma_\Delta$, where $\sigma_{\Delta,j}=\max(1.4826\,\mathrm{MAD}_j,0.25\,\mathrm{std}_j,10^{-3})$ is computed over the normalized demonstration displacements; it is then concatenated with a learnable gripper embedding $e(g_i)$, and finally mapped to a residual $r$ by an MLP whose output projection is zero-initialized. $r$ is added to the AdaRMS condition of every layer of the expert, so that the goal can reach every layer of the network, thereby strengthening the influence of the waypoint on action generation. At initialization $r=0$, and the expert behaves the same as before, so the original weights are not disrupted. Table~\ref{tab:faith} compares the conditioning variants.

If no constraint is placed on this route (which we call naive goal injection), it over-amplifies the influence of the waypoint. In the demonstrations, the relative goal is not an independent condition: $q_i-s_i$ approximately equals the integral of the very actions the expert is learning to produce, so delivering it to every layer amounts to leaking the label into the condition. The expert can then fit the training loss by integrating the displacement, without having to look at the images. We call this the displacement-integrator shortcut; Section~\ref{sec:faith} shows that it yields a very large $S$ but a low success rate.

NGM keeps the deep route and constrains it to discourage this shortcut. The goal is useful for the rough direction of a segment, whereas the final refinement should come from what the expert sees, so we use a fixed phase gate to restrict the deep route to the coarse stage of denoising:
\begin{equation}
\mathrm{cond}_\ell=\phi(t)+g(t)r,\qquad g(t)=\operatorname{clip}\left(\frac{t-0.2}{0.3},0,1\right).
\label{eq:ngm}
\end{equation}

The shortcut also requires the condition to correspond exactly to the label, so during training we add Gaussian noise to the endpoint ($\sigma_c=0.7$ in the normalized $\widehat\Delta$ domain) while the action supervision remains unchanged; in this way the goal can only serve as an approximate target, and the expert must cross-check it against the images. Finally, with probability 0.15 we replace the goal with a learnable null embedding, so that the expert remains capable without a goal. Both perturbations are applied to the suffix tokens and the deep route simultaneously; otherwise, an unperturbed suffix-token endpoint would bypass the regularization and bring the label in unchanged. At inference, noise and dropout are turned off while the gate remains on; the reported results use the ordinary conditional branch and do not use classifier-free guidance~\citep{cfg}.

\subsection{Training}
\label{sec:training}

The three waypoint variants use the same data, initialization (the released \pio\ weights), random seed, and number of training steps. The backbone receives $\nabla\mathcal{L}_{\text{CE}}+\lambda\,\nabla\mathcal{L}_{\text{FM}}$, where $\lambda=1.7$. Knowledge insulation blocks action-expert gradients from reaching the backbone~\citep{ki}; our variant retains this gradient path with weight $\lambda=1.7$. We apply LoRA~\citep{lora} (rank 16) to all linear layers in the backbone, the vision encoder, and the action expert, giving 46.4M--49.6M trainable parameters, or 1.27--1.35\% of the 3.66B total. The global batch for training the LIBERO models contains 80 planner windows and 80 action-expert segments.

\begin{table}[t]
\caption{Planning cost (A) and waypoint statistics (B).}
\label{tab:cost}
\centering\footnotesize
\setlength{\tabcolsep}{4pt}
\begin{minipage}[t]{0.43\textwidth}
\textit{A. Planning cost}\par\vspace{3pt}
\begin{tabularx}{\linewidth}{@{}>{\raggedright\arraybackslash}X cc@{}}
\toprule
 & Token-AR & Block-AR \\
\midrule
Serial passes / plan, max (LIBERO / robot) & 57 / 120 & 8 / 8 \\
\addlinespace[3.28pt]
Plan latency, robot (ms) & 1094 & 125 \\
\addlinespace[3.28pt]
Request round trip, robot (ms) & 1266 & 293 \\
\addlinespace[3.28pt]
Motion duty cycle, robot & 28.7\% & 65.6\% \\
\bottomrule
\end{tabularx}
\end{minipage}\hfill
\begin{minipage}[t]{0.53\textwidth}
\textit{B. Waypoint statistics and end-of-plan code}\par\vspace{3pt}
\begin{tabularx}{\linewidth}{@{}>{\raggedright\arraybackslash}X cc@{}}
\toprule
 & LIBERO-Long & robot \\
\midrule
Segment length p50 / p95 (steps) & 7 / 13 & 12 / 32 \\
Segments at the cap $D=32$ & 0\% & 5.8\% \\
\midrule
 & Token-AR & Block-AR \\
\cmidrule(l){2-3}
End-of-plan recall / precision & 0.00 / -- & 0.61 / 0.85 \\
Decoded plans with end marker, robot & 0 / 497 & 109 / 936 \\
\bottomrule
\end{tabularx}
\end{minipage}
\tabnote{\textwidth}{A: one prefix prefill and at most $M=7$ decoded blocks. Robot timings are means over the 497 Token-AR / 936 Block-AR requests of 35 telemetry episodes; expert and transport add about 160 and 7\,ms per request. Duty cycle: fraction of wall-clock time executing actions, averaged over episodes. B: 13{,}815 LIBERO-Long and 6{,}158 robot segments after extraction. End-of-plan recall/precision: 2{,}172 held-out robot windows, 270 true endings; Block-AR has 164 hits and 30 false alarms. The last row counts online plans containing an end marker. No LIBERO evaluation decoded a duration above $D$.}
\end{table}

\section{Experimental Setup}
\label{sec:setup}

\textbf{Simulation.} We conduct experiments on the four LIBERO suites~\citep{libero} (Spatial, Object, Goal, and Long). We train one model on the union of the four suites (1{,}693 episodes); waypoints are extracted with $\eta=0.008$ in the six-dimensional end-effector configuration, and waypoint selection also takes the open/close state of the gripper into account. For evaluation, each suite contains 10 tasks, and each task is evaluated on the 50 official initial states, i.e., $N=500$ episodes per suite; we follow the OpenPI LIBERO evaluation protocol~\citep{openpi_libero}, with step budgets of 220/280/300/520 steps, respectively, plus 10 settling steps. All evaluations start from the measured state, and LoRA weights are evaluated without exponential moving average (EMA).

\begin{figure}[!t]
\centering
\includegraphics[width=\textwidth]{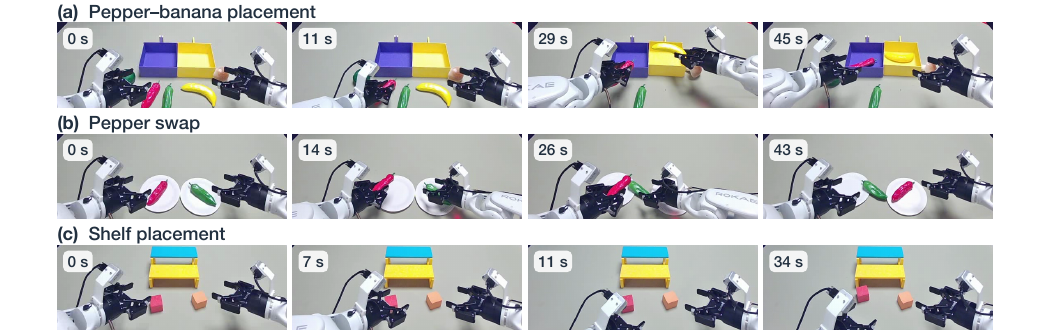}
\caption{Representative dual-arm sequences from Block-AR telemetry rollouts (external camera, seconds since start); success rates are reported separately in Table~\ref{tab:robot}. (a)~Pepper--banana placement. (b)~Pepper swap. (c)~Shelf placement, stopped at 34\,s with the red block carried to the shelf but not placed.}
\label{fig:robot}
\end{figure}

\textbf{Real robot.} The dual-arm platform consists of a pair of Rokae AR5-5 arms with Robotiq 2F-85 grippers; each arm is equipped with a wrist camera, and there is one additional external camera. We set up three bimanual tasks: placing a pepper and a banana into two boxes in the presence of distractors; swapping a red and a green pepper between two plates; and color-matched shelf placement (Fig.~\ref{fig:robot}). We collected 154 demonstrations in total (51, 51, and 52 for the three tasks); after excluding four episodes with inconsistent video length and holding out 15 for validation and 15 for testing, the remaining 120 (38, 40, and 42 for the three tasks) are used for training. Waypoints are extracted on the 14 joints with $\eta=0.015$\,rad. Each method is run for 20 trials per task in alternating order, with success judged by an operator.

\begin{table}[!t]
\caption{Waypoint dependence and visual responsiveness on LIBERO-Long.}
\label{tab:faith}
\centering\footnotesize
\setlength{\tabcolsep}{4pt}
\begin{tabular*}{\textwidth}{@{\extracolsep{\fill}}l c c c c c c c@{}}
\toprule
System & $B$ & $S_{\text{endpoint}}$ (\%) & $S_{\text{duration}}$ (\%) & $S_{\text{image}}$ & Retention & SR$_{\text{intact}}$ (\%) & $U$ (pp) \\
\midrule
Block-AR + suffix & 0.118 & 0.6 & 19 & 0.330 & 100\% & 91.0 & 0.0 \\
Token-AR + naive & 0.074 & 348 & 106 & 0.090 & 27\% & 15 & +14 \\
Block-AR + \NGM & 0.105 & 42.6 & 26.4 & 0.27 & 82\% & 96.2 & +7.4 \\
\bottomrule
\end{tabular*}
\tabnote{\textwidth}{All variants use the shared four-suite training and evaluation protocol. Naive injection adds the deep goal route without gate, condition noise, or dropout. Each intervention changes one input while holding the others and the flow noise fixed; $S$ and $B$ use the same 64 frozen segments. $B$: RMS action change under re-sampling the model's own flow noise, the denominator of $S$. $S$: median action change under endpoint erasure or duration $+4$ steps, relative to $B$. $S_{\text{image}}$: absolute RMS change under a main-camera swap; retention: its ratio to the suffix-token row. SR$_{\text{intact}}$: success with the unmodified plan; $U$: its drop after endpoint erasure on the same $N=500$ initial states per model.}
\end{table}

\section{Results}

\subsection{End-to-end performance}
\label{sec:e2e}

Table~\ref{tab:main} compares the results of the systems on the LIBERO dataset. Waypoint Token-AR differs from the reported $\pi_{0.5}$ results by at most 0.8 points on each suite ($-0.15$ points on average). Replacing Token-AR with Block-AR changes success by between $+0.6$ points (Spatial) and $-1.4$ points (Object, Goal), $-0.85$ points on average relative to Token-AR, while the number of forward passes of the VLM planner per plan drops from 57 to 8. With NGM, the same Block-AR system improves on every suite, by $2.6$ points on average and by $5.2$ points on LIBERO-Long. For any suite, one point corresponds to 5 of its 500 evaluation episodes; in addition, each variant was trained only once.

\subsection{Fast: the cost of a plan}
\label{sec:fast}

Table~\ref{tab:cost} and Fig.~\ref{fig:fast} quantify requirement (R1). On the real robot, each waypoint carries 19 tokens; Token-AR takes 1094\,ms to generate a plan and 1266\,ms for the full request including the action expert, longer than the roughly 0.5\,s action segment it commands, so the arms are in motion only 28.7\% of the time. Block-AR shortens the planning time to 125\,ms ($8.7\times$) and the round-trip latency to 293\,ms, and raises the motion duty cycle to 65.6\% (per task in Table~\ref{tab:robot}). The remaining latency comes mainly from the action expert (160\,ms per segment) rather than the planner. Among 2{,}172 held-out robot planning windows with 270 true plan endings, Token-AR never emits $d=0$, whereas Block-AR achieves 61\% recall and 85\% precision.

\subsection{Faithful: does the executor follow the plan?}
\label{sec:faith}

\textbf{Diagnosis.} Table~\ref{tab:faith} quantifies the extent to which the method with suffix-token conditioning satisfies requirement R2. On the same 64 action segments, with all other inputs and the flow-matching noise held fixed, the change in actions caused by erasing the planned endpoint amounts to only 0.6\% of the change caused by re-sampling the noise. However, the action expert is not generally insensitive to all of its inputs: swapping the main camera image changes the actions by 281\% of the noise re-sampling baseline; swapping the language instruction changes them by 158\%; and increasing the waypoint duration by 4 control steps changes them by 19\% of the noise baseline. In closed-loop evaluation, erasing every spatial endpoint in the plan causes no drop in task success, i.e., $U=0.0$.

\textbf{Sensitivity alone is insufficient to show that conditioning is effective.} Naive goal injection raises endpoint sensitivity to 348\% and duration sensitivity to 106\%. However, the model's response to visual changes falls to only 27\% of the suffix-token-conditioned baseline, and task success drops sharply to 15\%. All of its failures are timeouts, and its behavior shows the following pattern: the model relies excessively on the goal displacement proposed by the planner and does not make sufficient use of visual information for online correction, which is the displacement-integrator shortcut of Section~\ref{sec:ngm}. Endpoint sensitivity must therefore be read together with the success rate under the intact plan, the task utility of the endpoint, and the retained response to other inputs.

\textbf{Normalized goal modulation.} With NGM in place of suffix-only conditioning, the success rate of the Block-AR system on LIBERO-Long in closed-loop evaluation rises from 91.0\% to 96.2\%. The endpoint sensitivity of the NGM model increases from 0.6\% to 42.6\%, and duration sensitivity from 19\% to 26.4\%; at the same time, the sensitivity of the action expert to image changes remains at 82\%. Endpoint erasure, which costs nothing under suffix-token conditioning, now reduces success by 7.4 points: the expert uses the waypoint while keeping most of its response to vision.

\begin{table}[t]
\caption{Real-robot results.}
\label{tab:robot}
\centering\footnotesize
\setlength{\tabcolsep}{4pt}
\begin{tabular*}{\textwidth}{@{\extracolsep{\fill}}l c c c c c@{}}
\toprule
Task & Token-AR & Block-AR & +\NGM & Plan (ms) & Duty (\%) \\
\midrule
Pepper--banana & 17/20 & 16/20 & 18/20 & 1144/129 & 31/62 \\
Pepper swap & 16/20 & 15/20 & 17/20 & 1055/123 & 30/68 \\
Shelf placement & 15/20 & 15/20 & 16/20 & 1085/122 & 23/65 \\
\midrule
All & 48/60 & 46/60 & 51/60 & 1094/125 & 29/66 \\
\bottomrule
\end{tabular*}
\tabnote{\textwidth}{Successes over 20 trials per task and method, judged by the operator. Plan latency (ms, Token-AR / Block-AR) is averaged over requests and motion duty cycle (\%) over episodes, from the 35 telemetry rollouts of Fig.~\ref{fig:fast}. The All row averages over all requests and episodes, not over the three task rows.}
\end{table}

\subsection{Real robot}
\label{sec:robot}

Table~\ref{tab:robot} and Fig.~\ref{fig:robot} present the setup and results of the dual-arm robot experiments. Telemetry provides the latency and motion duty cycle: on every task, Block-AR reduces per-plan latency to $1/8.6$--$1/8.9$ of the original and more than doubles the fraction of time the arms are in motion. Each method is run for 60 trials; Token-AR, Block-AR, and Block-AR + NGM succeed 48, 46, and 51 times, respectively. On real-robot evaluation data not used in training, Block-AR also predicts the next waypoint more accurately: on 2{,}172 held-out planning windows, with the first-waypoint joint-angle error averaged over both arms, Token-AR has an error of 0.156\,rad and Block-AR 0.112\,rad, a 28\% reduction for the latter.

\section{Conclusion}

Our experiments show that end-to-end task success alone establishes neither planning efficiency nor effective use of the planner--executor interface in hierarchical VLAs. In a controlled Waypoint Token-AR hierarchy derived from $\pi_{0.5}$, we identify two forms of planner--executor misalignment: planning has a high inference latency, and the planned spatial endpoint barely contributes to action generation. Waypoint-aligned Block-AR keeps success within 1.4 points of Token-AR on every LIBERO suite while reducing serial planner passes from 57 to 8. NGM improves the Block-AR system on all four suites, by 2.6 points on average, and makes the executor depend more strongly on the planned endpoint: endpoint sensitivity rises from 0.6\% to 42.6\%, and erasing the endpoint reduces success by 7.4 points.

\textbf{Scope and limitations.} Block-AR and NGM are designed for and validated on one hierarchy derived from $\pi_{0.5}$: Block-AR assumes fixed-format plan token groups, and NGM relies on the AdaRMS conditioning of the $\pi_{0.5}$ expert. Whether they transfer to other VLA backbones is untested. Each model variant was trained only once, so we cannot estimate the variability across training runs. The real-robot study covers three tasks with 20 trials per task and method; this sample size limits the precision of success-rate estimates and conclusions about small between-method differences.

\clearpage
\newpage
\bibliographystyle{assets/plainnat}
\bibliography{refs}

\end{document}